# Importance of Sources using the Repeated Fusion Method and the Proportional Conflict Redistribution Rules #5 and #6

Florentin Smarandache
Math & Sciences Department
University of New Mexico, Gallup Campus, USA

Jean Dezert
French Aerospace Research Lab.
ONERA/DTIM/SIF
29 Avenue de la Division Leclerc
92320 Châtillon, France

**Abstract.**
We present in this paper some examples of how to compute by hand the PCR5 fusion rule for three sources, so the reader will better understand its mechanism.
We also take into consideration the importance of sources, which is different from the classical discounting of sources.

## 1. Introduction.

Discounting of Sources.
Discounting a source $m_1(.)$ with the coefficient $0 \le \alpha \le 1$ and a source $m_2(.)$ with a coefficient $0 \le \beta \le 1$ (because we are not very confident in them), means to adjust them to $m_1'(.)$ and $m_2'(.)$ such that:
$m_1'(A) = \alpha \cdot m_1(A)$ for $A \neq \Theta$ (total ignorance), and $m_1'(\Theta) = \alpha \cdot m_1(\Theta) + 1 - \alpha$,
*and* $m_2'(A) = \beta \cdot m_2(A)$ for $A \neq \Theta$ (total ignorance), and $m_2'(\Theta) = \beta \cdot m_2(\Theta) + 1 - \beta$.

Importance of Sources using Repeated Fusion.
But if a source is more important than another one (since a such source comes from a more important person with a decision power, let's say an executive director), for example if source $m2(.)$ is twice more important than source $m_1(.)$, then we can combine $m_1(.)$ with $m_2(.)$ and with $m_2(.)$, so we repeated $m_2(.)$ twice. Doing this procedure, the source which is repeated (combined) more times than another source attracts the result towards its masses – see an example below.
Jean Dezert has criticized this method since if a source is repeated say 4 times and other source is repeated *6* times, then combining *4* times $m_1(.)$ with 6 times $m_2(.)$ will give a result different from combining *2* times $m_1(.)$ with *3* times $m_2(.)$, although $4/6 = 2/3$. In order to avoid this, we take the simplified fraction $n/p$, where $gcd(n, p) = 1$, where *gcd* is the greatest common divisor of the natural numbers *n* and *p*.
This method is still controversial since after a large number of combining *n* times $m_1(.)$ with *p* times $m_2(.)$ for $n+p$ sufficiently large, the result is not much different from a previous one which combines $n_1$ times $m_1(.)$ with $p_1$ times $m_2(.)$ for $n_1+p_1$ sufficiently large but a little less than $n+p$, so the method is not well responding for large numbers.

A more efficacy method of importance of sources consists in taking into consideration the discounting on the empty set and then the normalization (see especially paper [1] and also[2]).

**2. Using $m_{PCR5}$ for 3 Sources.**

**Example** calculated by hand for combining three sources using *PCR5* fusion rule.

Let's say that $m_2(.)$ is 2 times more important than $m_1(.)$; therefore we fusion *m₁(.), m₂(.), m₂(.)*.

| | A | B | $A\cup B$ | $A\cap B=\Phi$ |
|---|---|---|---|---|
| $m_1$ | 0.1 | 0.7 | 0.2 | |
| $m_2$ | 0.4 | 0.1 | 0.5 | |
| $m_2$ | 0.4 | 0.1 | 0.5 | |
| $m_{122}$ | 0.193 | 0.274 | 0.050 | 0.483 |

$$\begin{cases} \frac{x_{1A}}{0.1}=\frac{y_{1B}}{0.7}=\frac{z_{2A\cup B}}{0.5}=\frac{0.005}{0.7}=\frac{0.05}{7} \\ x_{1A}=0.000714 \\ y_{1B}=0.000714 \\ z_{2A\cup B}=0.003572 \end{cases}$$

$$\begin{cases} \frac{x_{2A}}{0.4}=\frac{y_{2B}}{0.7}=\frac{z_{2A\cup B}}{0.5}=\frac{0.14}{1.6}=\frac{0.07}{0.8}=\frac{0.7}{8} \\ x_{2A}=0.035000 \\ y_{2B}=0.061250 \\ z_{2A\cup B}=0.043750 \end{cases}$$

$$\begin{cases} \frac{x_{3A}}{0.4}=\frac{y_{3B}}{0.1}=\frac{z_{3A\cup B}}{0.2}=\frac{0.008}{0.7}=\frac{0.08}{7} \\ x_{3A}\cong 0.004571 \\ y_{3B}\cong 0.001143 \\ z_{3A\cup B}\cong 0.002286 \end{cases}$$

$$\begin{cases} \frac{x_{4A}}{0.4}=\frac{y_{4B}}{0.1}=\frac{z_{4A\cup B}}{0.2}=\frac{(0.4)(0.1)(0.2)}{0.7}=\frac{0.008}{0.7}=\frac{0.08}{7} \\ x_{4A}\cong 0.004571 \\ y_{4B}\cong 0.001143 \\ z_{4A\cup B}\cong 0.002286 \end{cases}$$

$$\left[\begin{array}{l} \dfrac{x_{5A}}{0.4} = \dfrac{y_{5B}}{0.7} = \dfrac{z_{5A\cup B}}{0.5} = \dfrac{0.14}{1.6} = \dfrac{1.4}{16} \\ x_{5A} \cong 0.035000 \\ y_{5B} \cong 0.061250 \\ z_{5A\cup B} \cong 0.043750 \end{array}\right.$$

$$\left[\begin{array}{l} \dfrac{x_{6A}}{0.1} = \dfrac{y_{6B}}{0.1} = \dfrac{z_{6A\cup B}}{0.5} = \dfrac{0.005}{0.7} = \dfrac{0.05}{7} \\ x_{6A} \cong 0.000714 \\ y_{6B} \cong 0.000714 \\ z_{6A\cup B} \cong 0.003572 \end{array}\right.$$

$$\left[\begin{array}{l} \dfrac{x_{7A}}{0.1} = \dfrac{y_{7B}}{(0.1)(0.1)} = \dfrac{(0.1)(0.1)(0.1)}{0.1+0.01} = \dfrac{0.001}{0.11} \\ x_{6A} \cong 0.000909 \\ y_{6B} \cong 0.000091 \end{array}\right.$$

$$\left[\begin{array}{l} \dfrac{x_{8A}}{0.4} = \dfrac{y_{8B}}{(0.7)(0.1)} = \dfrac{(0.4)(0.7)(0.1)}{0.1+0.01} = \dfrac{0.028}{0.47} = \dfrac{2.8}{47} \\ x_{8A} \cong 0.023830 \\ y_{8B} \cong 0.004170 \end{array}\right.$$

$$x_{9A} = x_{8A} \cong 0.023830$$
$$y_{9B} = y_{8B} \cong 0.004170$$

$$\left[\begin{array}{l} \dfrac{x_{10A}}{(0.1)(0.4)} = \dfrac{y_{10B}}{0.1} = \dfrac{(0.1)(0.4)(0.1)}{0.04+0.1} = \dfrac{0.004}{0.14} = \dfrac{0.4}{14} = \dfrac{0.2}{7} \\ x_{10A} \cong 0.001143 \\ y_{8B} \cong 0.002857 \end{array}\right.$$

$$x_{11A} = x_{10A} \cong 0.001143$$
$$y_{11B} = y_{10B} \cong 0.002857$$

$$\begin{cases} \frac{x_{12A}}{(0.1)(0.4)} = \frac{y_{12B}}{0.1} = \frac{(0.4)(0.4)(0.7)}{0.16+0.7} = \frac{0.112}{0.86} = \frac{11.2}{86} \\ x_{12A} \cong 0.020837 \\ y_{12B} \cong 0.091163 \end{cases}$$

| | A | B | A⋃B |
|---|---|---|---|
| $m_{122}^{PCR5}$ | 0.345262 | 0.505522 | 0.149216 |

If we didn't double $m_2(.)$ in the fusion rule, we'd get a different result.
Let's suppose we only fusion $m_1(.)$ with $m_2(.)$:

| | A | B | A⋃B | A⋂B=Φ |
|---|---|---|---|---|
| $m_1$ | 0.1 | 0.7 | 0.2 | |
| $m_2$ | 0.4 | 0.1 | 0.5 | |
| $m_{12}$ | 0.17 | 0.44 | 0.10 | 0.29 |
| $m_{12}^{PCR5}$ | 0.322 | 0.668 | 0.100 | 0 |

And now we compare the fusion results:

| | A | B | A⋃B | |
|---|---|---|---|---|
| $m_{122}^{PCR5}$ | 0.345 | 0.506 | 0.149 | - *three sources*(*second – source – doubled*)*; importance of sources considered;* |
| $m_{12}^{PCR5}$ | 0.322 | 0.668 | 0.100 | - *two sources; importance of sources not considered*. |

The more times we repeat $m_2(.)$ the closer $m_{12...2}^{PCR5}(A) \to m_2(A)=0.4$, $m_{12...2}^{PCR5}(B) \to m_2(B)=0.1$, and $m_{12...2}^{PCR5}(A \cup B) \to m_2(A \cup B)=0.5$. Therefore, doubling, tripling, etc. a source, the mass of each element in the frame of discernment tends towards the mass value of that element in the repeated source (since that source is considered to have more importance than the others).

For the readers who want to do the previous calculation with a computer, here it is the $m_{PCR5}$
**Formula for 3 Sources:**

$$m_{PCR5}(A) = m_{123} + \sum_{\substack{X,Y \in G^{\Theta} \\ A \neq X \neq Y \neq A \\ A \cap X \cap Y = \Phi}} \left( \frac{m_1(A)^2 m_2(X) m_3(Y)}{m_1(A) + m_2(X) + m_3(Y)} + \right.$$

$$\left. + \frac{m_1(Y) m_2(A)^2 m_3(X)}{m_1(Y) + m_2(A) + m_3(X)} + \frac{m_1(X) m_2(Y) m_3(A)^2}{m_1(X) + m_2(Y) + m_3(A)} \right) +$$

$$+\sum_{\substack{X\in G^{\Theta}\\ A\cap X=\Phi}}\left(\frac{m_1(A)^2 m_2(X)m_3(X)}{m_1(A)+m_2(X)+m_3(X)}+\frac{m_1(X)m_2(A)^2 m_3(X)}{m_1(X)+m_2(A)+m_3(X)}+\frac{m_1(X)m_2(X)m_3(A)^2}{m_1(X)+m_2(X)+m_3(A)}\right)+$$

$$+\sum_{\substack{X\in G^{\Theta}\\ A\cap X=\Phi}}\left(\frac{m_1(A)^2 m_2(A)^2 m_3(X)}{m_1(A)+m_2(A)+m_3(X)}+\frac{m_1(X)m_2(A)^2 m_3(A)^2}{m_1(X)+m_2(A)+m_3(A)}+\frac{m_1(A)^2 m_2(X)m_3(A)^2}{m_1(A)+m_2(X)+m_3(A)}\right)$$

**3. Similarly, let's see the $m_{PCR6}$ Formula for 3 Sources:**

$$m_{PCR6}(A)=m_{123}+\sum_{\substack{X,Y\in G^{\Theta}\\ A\neq X\neq Y\neq A\\ A\cap X\cap Y=\Phi}}\left(\frac{m_1(A)^2 m_2(X)m_3(Y)}{m_1(A)+m_2(X)+m_3(Y)}+\right.$$

$$\left.+\frac{m_1(Y)m_2(A)^2 m_3(X)}{m_1(Y)+m_2(A)+m_3(X)}+\frac{m_1(X)m_2(Y)m_3(A)^2}{m_1(X)+m_2(Y)+m_3(A)}\right)+$$

$$+\sum_{\substack{X\in G^{\Theta}\\ A\cap X=\Phi}}\left(\frac{m_1(A)^2 m_2(X)m_3(X)}{m_1(A)+m_2(X)+m_3(X)}+\frac{m_1(X)m_2(A)^2 m_3(X)}{m_1(X)+m_2(A)+m_3(X)}+\frac{m_1(X)m_2(X)m_3(A)^2}{m_1(X)+m_2(X)+m_3(A)}\right)+$$

$$+\sum_{\substack{X\in G^{\Theta}\\ A\cap X=\Phi}}\left(\frac{m_1(A)^2 m_2(A)m_3(X)+m_1(A)m_2(A)^2 m_3(X)}{m_1(A)+m_2(A)+m_3(X)}+\right.$$

$$+\frac{m_1(X)m_2(A)^2 m_3(A)+m_1(X)m_2(A)m_3(A)^2}{m_1(X)+m_2(A)+m_3(A)}+$$

$$\left.+\frac{m_1(A)^2 m_2(X)m_3(A)+m_1(A)m_2(X)m_3(A)^2}{m_1(A)+m_2(X)+m_3(A)}\right)$$

**4. A General Formula for $PCR6$ for $s\geq 2$ Sources.**

$$m_{PCR6}(A)=m_{12\ldots s}+\sum_{\substack{X_1,X_2,\ldots,X_{s-1}\in G^{\Theta}\\ X_i\neq A,\, i\in\{1,2,\ldots,s-1\}\\ \left(\bigcap_{i=1}^{s-1}X_i\right)\cap A=\Phi}}\sum_{k=1}^{s-1}\sum_{(i_1,i_2,\ldots,i_s)\in P(1,2,\ldots,s)}\left[m_{i_1}(A)+m_{i_2}(A)+\ldots+m_{i_k}(A)\right]\cdot$$

$$\cdot\frac{m_{i_1}(A)m_{i_2}(A)\ldots m_{i_k}(A)m_{i_{k+1}}(X_1)\ldots m_{i_s}(X_{s-k})}{m_{i_1}(A)+m_{i_2}(A)+\ldots+m_{i_k}(A)+m_{i_{k+1}}(X_1)+\ldots+m_{i_s}(X_{s-k})}$$

where *P(1, 2, ..., s)* is the set of all permutations of the elements *{1, 2, ..., s}*.

It should be observed that $X_1, X_2, \ldots, X_{s-1}$ may be different from each other, or some of them equal and others different, etc.

We wrote this *PCR6* general formula in the style of *PCR5*, different from Arnaud Martin & Christophe Oswald's notations, but actually doing the same thing. In order not to complicate the formula of *PCR6*, we did not use more summations or products after the third Sigma.

As a particular case:

$$m_{PCR6}(A) = m_{123} + \sum_{\substack{X_1, X_2 \in G^\Theta \\ X_1 \neq A, X_2 \neq A \\ X_1 \cap X_1 \cap A = \Phi}} \sum_{k=1}^{2} \sum_{(i_1,i_2,i_3)\in P(1,2,3)} \frac{\left[m_{i_1}(A) + \ldots + m_{i_k}(A)\right] m_{i_1}(A)\ldots m_{i_k}(A) m_{i_{k+1}}(X_1)\ldots m_{i_3}(X_2)}{m_{i_1}(A) + \ldots + m_{i_k}(A) + m_{i_{k+1}}(X_1) + \ldots + m_{i_3}(X_2)}$$

where $P(1,2,3)$ is the set of permutations of the elements $\{1,2,3\}$.

It should also be observed that $X_1$ may be different from or equal to $X_2$.

**Conclusion.**

The aim of this paper was to show how to manually compute *PCR5* for *3* sources on some examples, thus better understanding its essence. And also how to take into consideration the *importance of sources* doing the Repeated Fusion Method. We did not present the Method of Discounting to the Empty Set in order to emphasize the importance of sources, which is better than the first one, since the second method was the main topic of paper [2].

We also presented the *PCR5* formula for *3* sources (a particular case when $n=3$), and the general formula for *PCR6* in a different way but yet equivalent to Martin-Oswald's *PCR6* formula.